\documentclass[sigconf]{acmart}

\usepackage{booktabs} 

\usepackage{xcolor}

\usepackage{framed,multirow}
\usepackage{makecell}
\usepackage{array}
\usepackage{graphicx}
\usepackage{amsfonts}
\usepackage{color}
\usepackage{subcaption}
\usepackage{amsmath}
\usepackage[super]{nth}
\usepackage{latexsym}
\usepackage{hyperref}

\setcopyright{acmcopyright}

\acmDOI{10.1145/3412841.3441884}

\acmISBN{978-1-4503-8104-8/21/03}

\acmConference[SAC '21]{The 36th ACM/SIGAPP Symposium on Applied Computing}{March 22--26, 2021}{Virtual Event, Republic of Korea}
\acmBooktitle{The 36th ACM/SIGAPP Symposium on Applied Computing (SAC '21), March 22--26, 2021, Virtual Event, Republic of Korea}
\acmYear{2021}
\copyrightyear{2021}
\acmPrice{15.00}

\begin{document}
\title{Natural vs Balanced Distribution in Deep Learning on \\Whole Slide Images for Cancer Detection}

\author{Ismat Ara Reshma}
\authornote{Corresponding author}
\affiliation{%
    \institution{IRIT, UMR5505 CNRS, Univ. de Toulouse}
    \streetaddress{118 Route de Narbonne}
    \city{Toulouse}
    \state{France}
    \postcode{31062}
}
\email{ismat-ara.reshma@irit.fr}

\author{Sylvain Cussat-Blanc}
\affiliation{%
    \institution{IRIT, ANITI, UMR5505 CNRS, Univ. de Toulouse}
    \streetaddress{118 Route de Narbonne}
    \city{Toulouse}
    \state{France}
    \postcode{31062}
}
\email{sylvain.cussat-blanc@irit.fr}
\author{Radu Tudor Ionescu}
\affiliation{%
    \institution{Univ. of Bucharest}
    \streetaddress{14 Academiei}
    \city{Bucharest}
    \state{Romania}
    \postcode{010014}
}
\email{raducu.ionescu@gmail.com}

\author{Hervé Luga}
\affiliation{%
    \institution{IRIT, UMR5505 CNRS, Univ. de Toulouse}
    \streetaddress{118 Route de Narbonne}
    \city{Toulouse}
    \state{France}
    \postcode{31062}
}
\email{herve.luga@irit.fr}

\author{Josiane Mothe} 
\affiliation{%
    \institution{IRIT, UMR5505 CNRS, Univ. de Toulouse}
    \streetaddress{118 Route de Narbonne}
    \city{Toulouse}
    \state{France}
    \postcode{31062}
}
\email{josiane.mothe@irit.fr}

\renewcommand{\shortauthors}{Reshma et al.}

\begin{abstract}
The class distribution of data is one of the factors that regulates the performance of machine learning models. However, investigations on the impact of different distributions available in the literature are very few, sometimes absent for domain-specific tasks. In this paper, we analyze the impact of natural and balanced distributions of the training set in deep learning (DL) models applied on histological images, also known as whole slide images (WSIs). WSIs are considered as the gold standard for cancer diagnosis. In recent years, researchers have turned their attention to DL models to automate and accelerate the diagnosis process. In the training of such DL models, filtering out the non-regions-of-interest from the WSIs and adopting an artificial distribution---usually a balanced distribution---is a common trend.  
In our analysis, we show that keeping the WSIs data in their usual distribution---which we call natural distribution---for DL training produces fewer false positives (FPs) with comparable false negatives (FNs) than the artificially-obtained balanced distribution.
We conduct an empirical comparative study with 10 random folds for each distribution, comparing the resulting average performance levels in terms of five different evaluation metrics. Experimental results show the effectiveness of the natural distribution over the balanced one across all the evaluation metrics.

\end{abstract}

%
%


\maketitle

\section{INTRODUCTION}

Histological images or whole slide images (WSIs) are the digital conversions of conventional histological glass slides containing tissue samples \cite{wsi_farahani2015whole,wsi_kumar2020whole}.
During the last two decades, the analysis of WSIs has become common practice for the clinical diagnosis of cancers by pathologists. However, the process of diagnosis has some well-known limitations, such as being very time consuming, laborious, and subjective to pathologists' expertise and fatigue. To address these limitations, researchers are investigating automatic cancer detection using machine learning models. Specifically, the huge success of deep learning models, such as convolutional neural networks (CNNs), in visual recognition \cite{LeCun-review, Krizhevsky-imageNet, Rawat-review} motivated researchers to explore their use in cancer detection in WSIs \cite{JAMA2017, Wang-CAMELYON16, Liu-GoogleAI}. 

However, the success of CNN models depends on different hyper-parameters settings. One of the most important hyper-parameters is the class distribution of the training data. In machine learning, the imbalanced data distribution has been shown to lead to inferior models, compared to the balanced distribution \cite{oversamplingDrawback-chawla2002smote,imbalancedData-khan2017cost}. This is likely to be the reason why the balanced distribution is the most adopted in deep learning state-of-the-art methods \cite{JAMA2017,Liu-GoogleAI,balancedDist-in-wsi-halicek2019head}, although \citet{prati_imbalance_Revisit2015}, for example, showed it is not optimal in all cases. Unfortunately, very few  comparative studies exist in the literature \cite{weiss2001effect_class_dist, weiss2003learning_class_dist, prati_imbalance_Revisit2015}, 
and these are mainly conducted on toy data sets. However, real data sets can be far different and more complex than the toy ones. Thus, there is no evidence that the balanced distribution will work on cancer WSIs. Consequently, being a special kind of image type, domain-specific studies are required for WSIs data. 

\begin{figure}[!t]
\centering
\includegraphics[width=.45\textwidth]{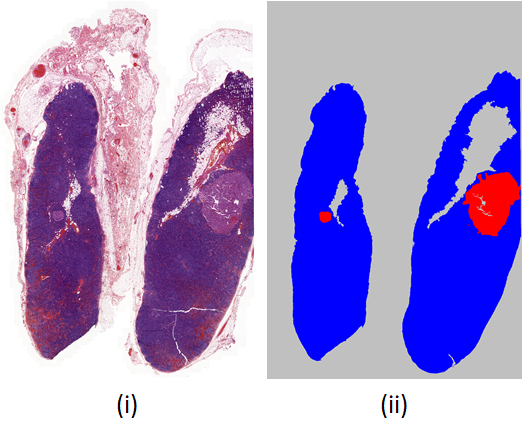}
\caption{A WSI of a metastatic lymph node (i) with its annotated mask (ii), where each color represents a specific part. The red regions represent metastasis (cancer) in the lymph node (ROI), the blue regions represent lymph node without metastasis (also ROI), while the gray regions represent background and other histological structures (non-ROI). Best viewed in color.} 
\label{fig:exampleOf_annotatedImg}
\end{figure}

In a whole slide image, some particular regions are considered as regions-of-interest (ROIs), drawing the interest of pathologists to check for abnormalities, and are in turn divided into several classes, depending on the use purpose. In contrast, the remaining regions are considered as non-ROIs, which are mainly background. 
For example, in a WSI of a metastatic lymph node, the regions comprising the lymph node tissue are ROIs, while the remaining ones are non-ROIs. In this case, ROIs are further divided into cancer, unhealthy versus non-cancer, healthy parts. Non-ROI parts are background as well as some other histological structures such as blood, adipose, or fibrous tissue. Fig.~\ref{fig:exampleOf_annotatedImg} presents an example WSI of metastatic lymph node with all the mentioned parts.

One of the important features in WSI data sets is that the pixel distribution is usually biased towards the non-ROI class, which we call the \textit{natural} distribution, that is to say, the distribution provided in the data set. 
Typically, a WSI contains 70 to 80\% of non-ROI pixels on average \cite{Wang-CAMELYON16, scannet-lin2018scannet}. It is common practice to remove the non-ROIs from the computation pipeline by applying some pixel filtering algorithms \cite{Wang-CAMELYON16, scannet-lin2018scannet}. However, to the best of our knowledge, the usability of non-ROIs during training has not been investigated so far.

Regarding the ROI classes, the natural distribution is either balanced or imbalanced depending on the included WSIs in the data set. It is common practice to artificially balance the ROI classes when they are not in the natural distribution of the data set. In the case of WSI, related work does not analyze the impact of these distributions on the results.

While related work concludes that it is usually more appropriate to balance the examples in each class in the training data set, we believe the cancer WSI case can be very different from the used toy data sets. Indeed, cancer WSI data have very specific inter-class similarities and intra-class differences. 
\citet{inter-intra-wang2015segmenting} consider this can lead to false positives (FPs). Although reducing false negatives (FNs)---thus getting high sensitivity---is also important in a biomedical task, reducing FPs requires more attention than FNs because of its difficulty level. Notably, it is easier to reduce FNs than FPs. For example, choosing a softer threshold for the predicted score can remove almost 100\% FNs, while reducing FPs with high sensitivity is not that straight forward. In this regard, over-representation of the error-causing class can reduce FPs without a significant reduction of sensitivity. In the case of WSIs, over-representation of non-ROIs, i.e., the natural distribution of the training set, can achieve the same effect. Indeed, according to our empirical observations, ROIs and non-ROIs contain some common histological structures, which we think could be useful in resolving the confusion regarding class variability and similarity during training. In other words, increasing the number of non-ROIs in the training set forces the network to classify these common structures as non-ROI, thus improving the final result. Consequently, filtering out non-ROI may not be the best practice in such cases.

In this paper, we present an empirical comparative study between the popular artificially-balanced and imbalanced distribution (e.g. original distribution of real data) of WSI data. 
Through this study, we also found useful insights regarding the usefulness of non-ROIs during training. We believe that the results of our study are specifically important when building new data sets, in order to know which class examples are worth annotating.

The rest of this paper is structured as follows. We first present the related work in Section~\ref{sec:related-work}.  
We describe our methodological framework along with the data set, hyper-parameter settings, and evaluation metrics in Section~\ref{sec:methods}, while the results are presented and discussed in Section~\ref{sec:result-discussion}. In Section~\ref{sec:conclusion}, we draw some conclusions and mention  future directions. 

\section{RELATED WORK}

\label{sec:related-work}

Most of the real-world data is skewed towards certain classes \cite{imbalancedData-roy2018study}. This skewness is usually considered as a problem that makes the classification task challenging for a classifier. Many studies have been carried out to measure the impact of artificially-balanced training sets, some suggesting straightforward data-level modifications \cite{oversampling-jaccard2017detection,oversampling-levi2015age,oversamplingDrawback-chawla2002smote,downsampling-kubat1997addressing}, such as oversampling, undersampling, or using synthetic data, and others suggesting algorithm-level modifications \cite{imbalancedData-khan2017cost,imbalanceData_sun2007cost}, such as applying a weighted cost function. There are even pieces of work that tried to combine data-level and algorithm-level changes \cite{hybrid-chawla2003smoteboost,imbalancedData-yuan2018regularizedEnsamble,imbalancedData-wu2020smoteEnsamble}. \citet{prati_imbalance_Revisit2015} listed some of the most popular data and model balancing methods, focusing on classical machine learning problems, yet, only a few studies discuss the deep learning perspective \cite{imbalancedData-Buda2017,imbalancedData-johnson2019survey}.
\citet{imbalancedData-Buda2017} compared different methods to address the class-biased problem. According to the authors,  imbalanced data has an adversarial effect on the classification accuracy of CNNs, as for classical machine learning techniques. 
\citet{imbalancedData-johnson2019survey} conducted a detailed survey on recent techniques to deal with the imbalanced data problem in deep learning, concluding that no method has enough evidence of superiority in dealing with class imbalance using deep learning.

All the aforementioned articles 
consider the imbalanced data as inauspicious for learning, suggesting methods to make the class distribution of the training data balanced, although the generalization of such methods to specific domains is yet to be proven. In particular, comparative studies on different distributions are required to make a conclusive decision on each data category. However, such studies are few.
\citet{weiss2001effect_class_dist, weiss2003learning_class_dist} showed that neither the naturally-occurring class distribution nor the balanced distribution is the best for learning, and often, substantially better performance can be obtained by using a different class distribution. In their analysis, they employed 26 data sets from the UCI repository \cite{UCI_2017}. According to the analysis of \citet{prati_imbalance_Revisit2015} on 20 data sets from UCI and a few other private data sets, the most recommended distribution for seven different learning algorithms, including neural networks, is the balanced distribution. 
They conclude that only Support Vector Machines are less affected by the class imbalance. However, the study was done on toy data sets.  
Thus, there is no evidence that the commonly-prescribed balanced distribution is a generalizable solution for more complex real data sets. Consequently, a separate dedicated analysis is required for every special kind of data, like WSIs. 

To our knowledge, there is no such comparative study for the cancer detection task from WSIs, while this is beneficial to be known for an optimal outcome of a cancer detection system. Moreover, in state-of-the-art cancer detection systems for WSIs \cite{Wang-CAMELYON16,Liu-GoogleAI,liu2018-googleAI_LYNA,cancerD-networkArchi-lin2018scannet,data-PCam-veeling2018rotation}, the common practice is to filter out non-ROIs and to create a training set with an artificially-balanced or a slightly-skewed distribution to the negative ROI class. Thus, the original data distribution and the use of non-ROIs are yet to be explored.

\section{Methodological framework }
\label{sec:methods}

\subsection{Different Class Distributions}
To evaluate the importance of the class distribution in the training set, we run a series of experiments that use two different distributions. 
Specifically, we consider the \textit{natural distribution}, which refers to the initial distribution of the data set, i.e., the distribution a WSI data set naturally has or as provided by stakeholders or pathologists.  
On the other hand, we consider the \textit{balanced distribution}, which is artificially generated such that classes become uniformly distributed.

WSIs are a special kind of image. They are produced by a whole slide scanner which is a microscope under robotic and computer control, equipped with highly specialized cameras containing advanced optical sensors \cite{wsi_farahani2015whole}. 
Unlike a usual image, a WSI contains information of multiple levels (usually, 8 to 10) of resolution in the same file that allows the extraction of images with low to high magnification levels. The extracted WSIs are usually very large in size, considering the level of extraction.  For example, at $40\times$ magnification or level 0, a WSI could be several gigapixels in size. It is impractical to use gigapixels WSIs as direct input to CNNs. 
To tackle the memory issue, the usual practice is to consider patches rather than whole WSIs for CNN training. We also follow this principle in our study. Thus, we extract overlapping patches and corresponding ground-truth annotation masks from the annotated training data of metastatic lymph node WSIs.

The extracted patches are categorized into different categories to facilitate the creation of different class distributions for the analysis. 
These categories of patches are defined according to the pixel classes of the patches. 
In particular, patches that contain more than 99.99\% non-ROI pixels are labeled as the \textit{other} ($\mathbb{O}$) category. The remaining patches belong to one of the following categories based on the presence of corresponding class pixels: \textit{cancer} ($\mathbb{C}$), \textit{lymph node} ($\lnot\mathbb{C}$), and \textit{mixed} ($\mathbb{C}\&\lnot\mathbb{C}$) with an optional presence of class $\mathbb{O}$ pixels. Among all the patch categories, we consider the category $\mathbb{C}\&\lnot\mathbb{C}$ as a multiclass patch category (since it contains two ROI classes at the same time).

\begin{table}
\caption{Experiments: balanced distribution (E.a) and natural over-representation of class $\mathbb{O}$ (E.b) with a total of 9 units of patches in the training set of each experiment.}  
\label{tab:exp-configH1}
\centering
\setlength{\tabcolsep}{8pt}
\begin{tabular}{lcc}
\toprule
  Experiment ID & Distribution & \vtop{\hbox{\strut Patch ratio}\hbox{\strut($\mathbb{O}:\mathbb{C}:\lnot\mathbb{C}$)}}\\
\midrule
  E.a & Balanced &	$3:3:3$ \\ 
  E.b & Natural&	$7:1:1$ \\  
\bottomrule
\end{tabular}
\end{table}

We generated the two distributions of classes in the training set by selecting patches from these different categories, as shown in Table~\ref{tab:exp-configH1}. 
To train a model with a balanced distribution, we design experiment E.a by considering the same number of patches from each of the three classes. On the other hand, to train a model with the natural distribution, we design E.b, where, the training set is highly biased (7 times) toward the class $\mathbb{O}$ similar to the usual natural distribution in the WSI. In other words, in the training set of E.b, almost 78\% of examples are from class $\mathbb{O}$. 
We keep the total number of patches equal in both experiment settings for the sake of fair comparison. Hence, it is not possible to utilize all the extracted patches at the same time in an experiment. Rather, a subset from each category is selected randomly by maintaining the class ratio of a distribution in a particular experiment.

\subsection{Network and Hyper-Parameter Setting} 

We selected U-net \cite{unet} as our CNN architecture because it has been proven to be effective even when using very few training images and its training time is manageable. We implemented the U-net architecture using Keras \cite{chollet2015keras} with the TensorFlow backend. 

In all the experiments, 80\% of the training data samples are used to train the model, the remaining 20\% being kept for validation. The validation set is selected randomly from the training set. The models were trained from scratch, i.e., without using transfer learning. We evaluated the trained model on the designated test data set, which was completely unseen during training.  

After preliminary empirical evaluation, we set the number of epochs to $35$ and the mini-batch size to $5$. We opted for the categorical cross-entropy as loss function, similar to the original U-net. To optimize the objective function, we used the Adam optimizer \cite{Kingma-ICLR-2015}. 
The initial weights were drawn randomly from the zero-mean Gaussian distribution, as recommended by \citet{Krizhevsky-imageNet}. We used a standard deviation of $0.05$, which is the default setting in Keras. 

We ran the same experiments for different learning rates ($10^{-5}$, $10^{-4}$) and different seeds, observing consistent results across the various configurations. Hence, in Section~\ref{sec:result-discussion}, we report the results using one setting only.

\subsection{Training / Testing Principle}
In the training step, we generate the natural and the balanced class distributions in the training set by selecting patches from the different categories as defined in the previous sub-section and considering the distribution in Table~\ref{tab:exp-configH1}. 

The random patch selection is followed by the shuffling of the whole training set to prevent all patches in a mini-batch being from the same category. This makes the convergence faster during training and provides better accuracy \cite{koller2015deepL}. Once the training set with a particular class distribution is created, the CNN model is trained on top of it. Since the CNN model is influenced by the random weight initialization and the stochastic training process, training the model once might not be enough to validate our hypothesis. Thus, we perform 10 runs for each class distribution, averaging the corresponding results. For two experiments presented in Table~\ref{tab:exp-configH1}, we have a total of $2\times10=20$ runs for a particular parameter setting.

During inference, the trained model is employed to predict the annotation mask of the unseen test WSIs.
For this purpose, equally-sized overlapping patches from the test WSIs are considered for prediction. We empirically set half of the pixels of a test patch in the middle as a central region (CR) and the remaining part as a border for each patch. Although the patches are overlapping, the CRs themselves are non-overlapping in a test WSI. The trained model predicts the pixel-level confidence score for each class in the corresponding patch. 
We annotate a pixel with the class corresponding to the highest confidence score. However, the predicted score for the CR of a patch is taken into account during evaluation.

\subsection{Data Set}  
\label{sec:dataset}

We utilized the Metastatic Lymph Node data set, named MLNTO, from the University Cancer Institute Toulouse Oncopole. 
MLNTO is a private data set containing metastases (cancer that spread from its origin to other parts of the body) of 16 primary cancer types and organs, for example, melanoma, adenocarcinoma and squamous cell carcinoma of various anatomical sites. 

The data set contains WSIs of lymph nodes stained with hematoxylin and eosin (H\&E). Two expert pathologists have provided the ground-truth segmentation masks for all of the WSIs. The three classes as mentioned previously were considered during the annotation (see Fig.~\ref{fig:exampleOf_annotatedImg}): \textit{metastasis/cancer} ($\mathbb{C}$), \textit{lymph node} ($\lnot\mathbb{C}$), and so-called \textit{other} ($\mathbb{O}$), the latter being either background or histological structures, such as adipose or fibrous tissue. The ground-truth masks are given for WSIs that are downsampled by a factor of 8 from the highest resolution, i.e., at level 3 resolution. 

We first extract patches from the WSIs. Although random patch sampling is usually applied for patch extraction \cite{Wang-CAMELYON16,Liu-GoogleAI,cancerD-networkArchi-lin2018scannet}, we extract overlapping patches by maintaining a fixed stride, since we empirically found that using a regular grid performs better than random sampling. Furthermore, there is a lack of sufficient contextual information at the edges of a patch, and CNNs, like U-net, mostly focus on the center region of an input image. Hence, it is preferable to learn from overlapping patches to capture all regions of a WSI.

\begin{table}
\caption{The number of patches that belong to each category in the training set of the MLNTO data set, when WSIs are downsampled by a factor of $8\times$ and the stride is $d/2$ (i.e., 192).}
\label{tab:t-patch_categ}
\centering
\begin{tabular}{lr}
\toprule
Patch category & \#patches \\
\midrule
 \textit{Other} ($\mathbb{O}$) & $90,374$ \\
 \textit{Cancer} ($\mathbb{C}$) & $15,328$  \\
 \textit{Lymph node} ($\lnot\mathbb{C}$) & $17,274$ \\
 \textit{Mixed} ($\mathbb{C}$\&$\lnot\mathbb{C}$) & $4,922$ \\
 \bottomrule
\end{tabular}
\end{table}

A total of 127,898 overlapping patches of dimension ($d\times d$ =) $384\times384$ pixels have been extracted with a stride of $d/2$ from the training set. The value of $d$ has been empirically determined for the segmentation task. The number of patches that fall into each category are given in Table~\ref{tab:t-patch_categ}. 

In our study, we exclude the patches from the mixed ($\mathbb{C}\&\lnot\mathbb{C}$) categories, since these might be problematic in loss calculation and gradient optimization for CNNs, according to \citet{excludeMixedPatches-halicek2019head}.
Among the remaining categories, $\mathbb{C}$ has the minimum number of patches (15k), thus, for a balanced distribution, we consider 15k patches from each category in E.a. Likewise, in E.b, to respect the natural data imbalance without increasing the total number of patches, we considered 35k patches from the category $\mathbb{O}$ and 5k from each of the other two categories, $\mathbb{C}$ and $\lnot\mathbb{C}$. In summary, the training data for each experiment consists of a total of 45k patches.

\begin{table}
\caption{A pixel statistics reporting the average number (in millions and percentage) of pixels of each class per WSI in the training and test sets of MLNTO data set. Considered WSIs are downsampled by a factor of from the original resolution.}
\label{tab:pixel-stat-train-test}
\centering
\begin{tabular}{l@{}l@{\hspace{.5em}}c@{\hspace{.5em}}c@{\hspace{.5em}}c@{\hspace{.5em}}l@{}}
\toprule
& & $\mathbb{C}$ & $\lnot\mathbb{C}$ & $\mathbb{O}$ \\
\midrule

\multirow{2}{1cm}{Train} 
& Mean in millions & 15.2 & 14.6 & 107.4  \\
& Mean in \% & 11.1 & 10.6 & 78.3 \\
\midrule

\multirow{2}{1cm}{Test}
& Mean in millions & 20.0 & 9.8 & 114.8\\
& Mean in \% & 13.8 & 6.8 & 79.4\\
\bottomrule

\end{tabular}
\end{table}

\begin{figure}[!h]
\centering
\includegraphics[width=.47\textwidth]{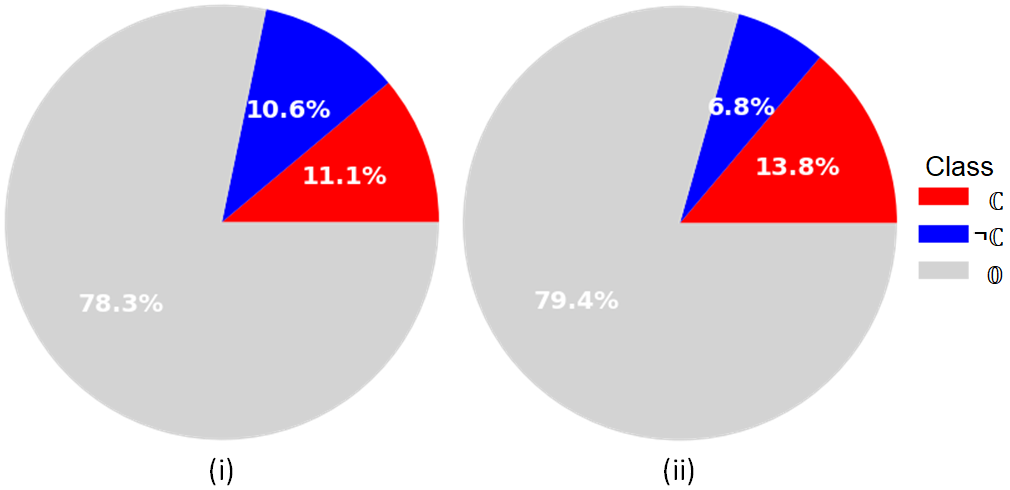}
\caption{Class distributions of the pixels in the training (i) and test set (ii) of the MLNTO data set.}
\label{fig:pixelStat}
\end{figure}

 Table~\ref{tab:pixel-stat-train-test} and Fig.~\ref{fig:pixelStat} show the class distributions of the pixels in the training and test data sets. The statistics show that the natural data distribution is imbalanced with an over-representation of the class $\mathbb{O}$ in comparison to classes $\mathbb{C}$ and $\lnot\mathbb{C}$.
 In the training set, the percentage of class $\mathbb{O}$, $\mathbb{C}$, and $\lnot\mathbb{C}$ in a WSI are on average 78.3\%, 11.1\%, and 10.6\%, respectively, while in the test set they are 79.4\%, 13.8\%, and 6.8\%, respectively. 
 The imbalanced nature of the data convinced us to better understand the impact of balanced versus imbalanced distribution of the classes on the trained model.

\subsection{Evaluation Metrics}
\label{sec:eval-metric}
For a clinical system, false negative (FN) and false positive (FP) values are both important. Therefore, we considered evaluation metrics that capture both features. 

In particular, we utilized the following well-known evaluation metrics to evaluate our trained models: accuracy (Acc), precision (P), a.k.a positive predictive value (PPV), recall (R), a.k.a sensitivity or true positive rate (TPR), F1-measure, 
receiver operating characteristic (ROC) curve, and precision-recall (PR) curve. Here, the former four metrics were computed based on the argmax decision (based on a single threshold or cutoff point) of the model's prediction scores, while the latter two metrics were computed based on all possible threshold values. The precision and recall are responsive to the FP and FN, respectively, while the PR curve expresses the trade-off between precision and recall. Another widely-used metric, the ROC curve, plots the recall/TPR at every false positive rate (FPR) point. Here, FPR is equivalent to 1-specificity. 

Furthermore, we applied F1-measure, accuracy, and area under curve (AUC) to summarize the performance of the models. Among all the metrics, the precision, F1-measure, and PR curve can capture the poor performance of the classifier for the imbalanced test set \cite{eval-saito2015precision}. 

All aforementioned metrics were applied in our patch-level evaluation since it is impractical to consider each pixel in a WSI. In this regard, we employed some post-processing on the predicted scores to compute the curve-based evaluation metrics. More precisely, we took the maximum of the predicted scores of all CR pixels in a test patch as the predicted score of that patch for a particular class, in this case, for class $\mathbb{C}$. To calculate the mean curve of the 10 runs, we normalized the predicted scores for each run by applying min-max normalization then we used interpolation to compute the mean curve.

\section{Results and discussions}

\label{sec:result-discussion}

\begin{figure}[!h]
\centering
\includegraphics[width=0.4\textwidth]{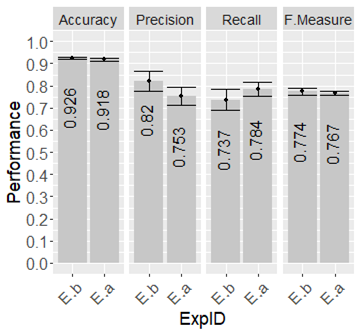}
\caption{Comparison between experiments, E.a (for balanced distribution) and E.b (for natural distribution) for class $\mathbb{C}$. Here, the top of each bar represents mean performance and the black line segment, which is known as the error bar, represents $\pm$ standard deviation. The experiments are arranged according to the ascending order of the precision.}
\label{fig:result-h1-barPlot-c-nc}
\end{figure}

Fig.~\ref{fig:result-h1-barPlot-c-nc} presents the results of our experiments based on the argmax decision on the predicted scores of the models. We present both the mean and standard deviation considering the 10 runs.

In Fig.~\ref{fig:result-h1-barPlot-c-nc}, the results show that the naturally imbalanced distribution (E.b) is better than the artificially balanced distribution (E.a) in terms of all evaluation metrics, except for recall. The precision of E.b (second sub-part) is 7\% higher than that of E.a., which means that the distribution with over-represented class $\mathbb{O}$ (E.b) produces less false positives than the balanced distribution (E.a). According to the t-test on the precision, the difference is statistically significant (p-value $<$ .002).

On the other hand, according to the recall (third part of Fig.~\ref{fig:result-h1-barPlot-c-nc}), the result is opposite, namely the balanced distribution (E.a) gives higher recall than the other distribution. One of the reasons could be the number of positive examples in the training set. To keep the same total number of training examples in both experiments (for a fair comparison), we did not use all the positive examples (class $\mathbb{C}$) in E.b, while in E.a, we utilized almost all positive examples. 

Since recall and precision are inversely proportional, we also plot the F1-measure {(which is the harmonic mean of the precision and recall)} to compare the overall result by considering both precision and recall at the same time. According to the F1-measure, E.b is better than the E.a. 

\begin{figure}[!h]
\centering
\includegraphics[width=.5\textwidth]{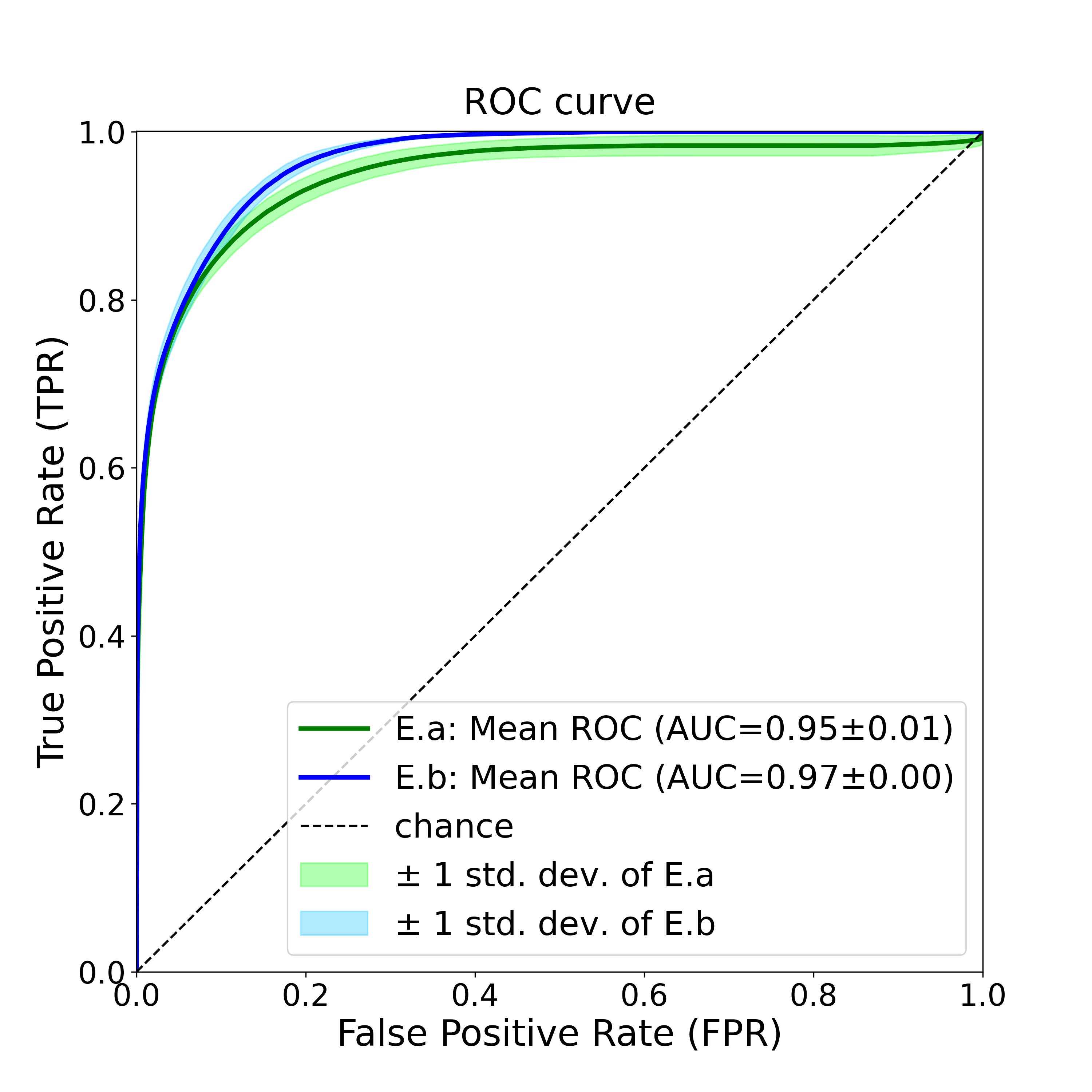}
\caption{Mean ROC curves with standard deviation (std. dev.) of E.a (balanced distribution) and E.b (natural distribution) for class $\mathbb{C}$.}
\label{fig:roc-curves}
\end{figure}

\begin{figure}[!h]
\centering
\includegraphics[width=.5\textwidth]{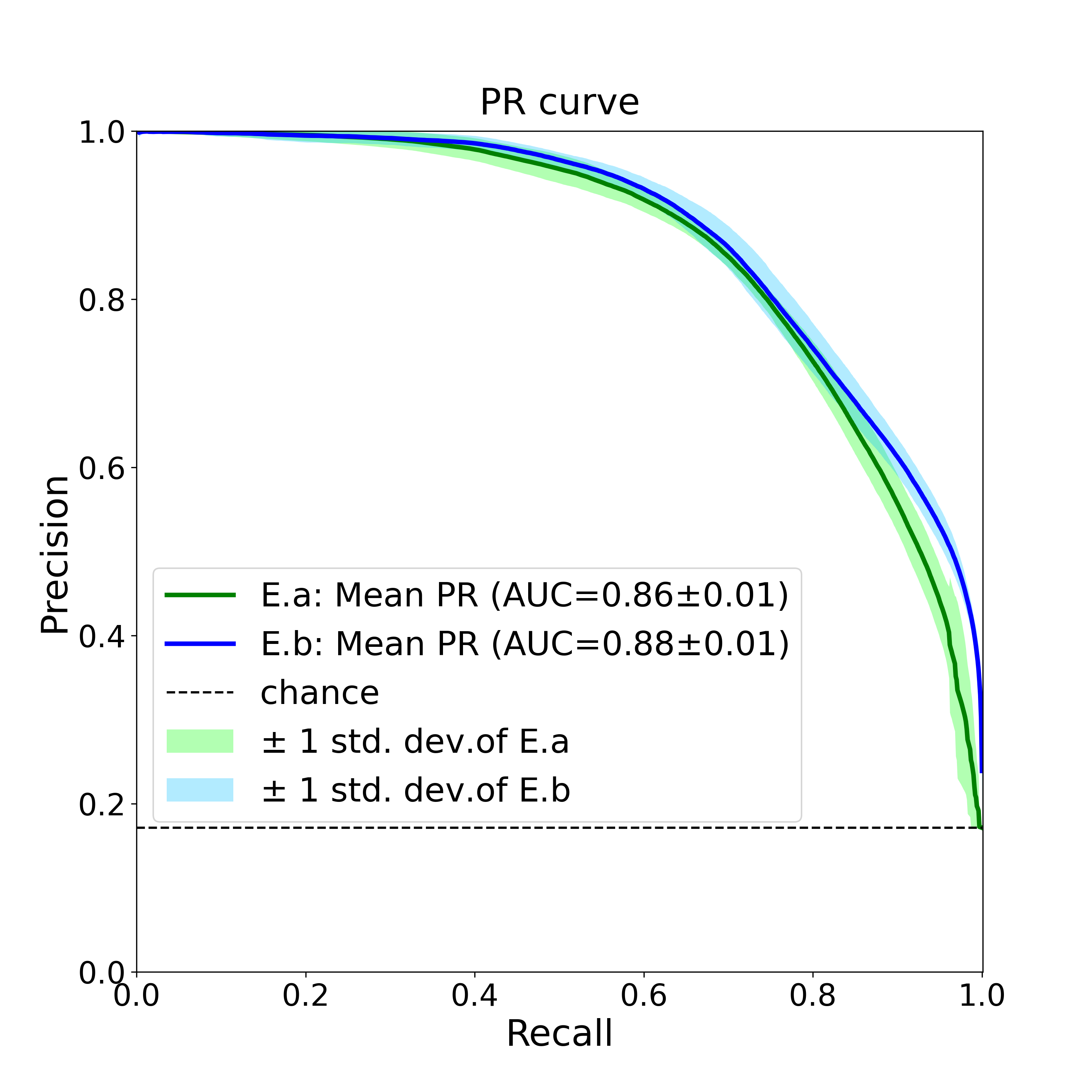}
\caption{Mean PR curves with standard deviation (std. dev.) of E.a (balanced distribution) and E.b (natural distribution) for class $\mathbb{C}$.}
\label{fig:pr-curves}
\end{figure}

While Fig.~\ref{fig:result-h1-barPlot-c-nc} presents results for a single threshold point (cutoff), Fig.~\ref{fig:roc-curves} and \ref{fig:pr-curves} illustrate the results for all possible threshold points. In these figures, we present the mean curves resulting from the 10 runs as well as the standard deviation.  

According to both ROC and PR curves, the natural distribution (E.b) is better than the balanced distribution (E.a). In Fig.~\ref{fig:roc-curves}, we can observe that E.b has higher TPR (i.e., sensitivity or recall) than E.a, at almost every FPR points. In other words, models trained under the natural distribution give higher recall, while also producing less false positives (FP) than the models trained under the balanced distribution. Moreover, the standard deviation of the E.b is less than that of E.a, and most of the time non-overlapping. 

Although the ROC curve is a popular metric to evaluate biomedical models, neither TPR nor FPR can capture the poor performance of the models for the imbalanced test data \cite{eval-saito2015precision}. Thus, as an alternative, we  report the results in terms of the PR curve as well (see Fig.~\ref{fig:pr-curves}). According to the PR curve, the precision (i.e., PPV) of E.b is higher at almost every recall (i.e., TPR/sensitivity) point than E.a. Even at the recall point 1.0 (i.e., at 100\% recall), the precision of E.b is better than that of the random chance baseline, while the precision of E.a is as low as the random chance. Furthermore, similar to the ROC curve, the standard deviation of E.b is less than that of E.a, and at high recall, they are non-overlapping.

In Table~\ref{tab:example-pred}, we present some example predictions of E.a ($3^{rd}$ row) and E.b ($4^{th}$ row) with the corresponding test WSIs ($1^{st}$ row) and ground-truth (GT) masks ($2^{nd}$ row). Here, we choose the WSIs with three characteristics, most FP generative ($1^{st}$ column), WSI with micro metastasis/$\mathbb{C}$ ($2^{nd}$ column), and full negative ($3^{rd}$ column). The predictions show that E.b produces less FP than E.a.

\begin{table}
\caption{Examples with ground-truth annotations and predictions taken from both E.a and E.b experiments. Best viewed in color.}
\centering
\begin{tabular}{@{}m{1em}p{19mm}p{19mm}p{19mm}@{}}

\toprule
& Most FP & Micro $\mathbb{C}$ & Full negative\\

\midrule
\rotatebox[origin=c]{270}{WSI} & 
\includegraphics[width=2.cm,height=3.4cm]{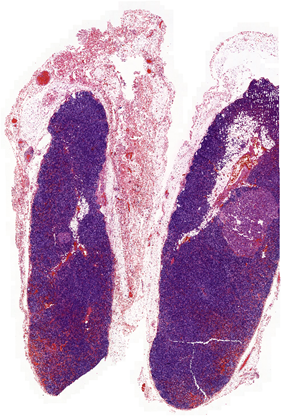} & \includegraphics[width=2.cm,height=3.4cm]{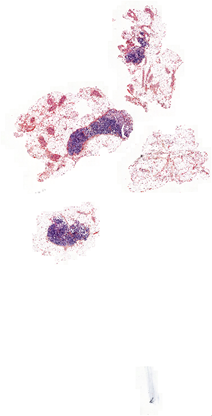} & \includegraphics[width=2.cm,height=3.4cm]{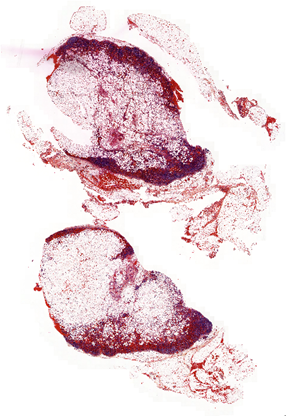} \\

\midrule
\rotatebox[origin=c]{270}{GT} &  
\includegraphics[width=2.cm,height=3.4cm]{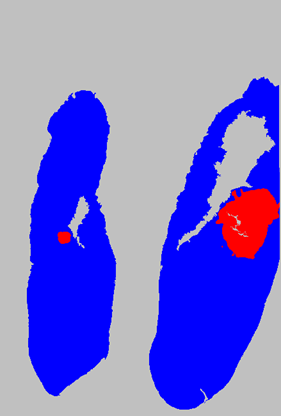} & \includegraphics[width=2.cm,height=3.4cm]{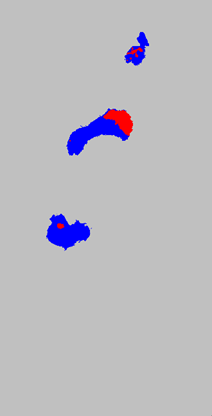} & \includegraphics[width=2.cm,height=3.4cm]{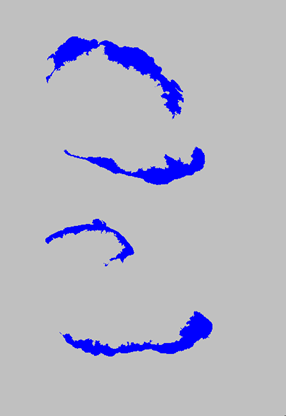}\\

\midrule
\rotatebox[origin=c]{270}{E.a} & 
\includegraphics[width=2.cm,height=3.4cm]{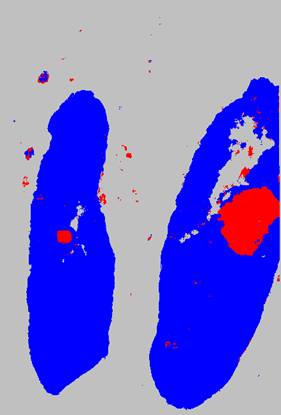} & \includegraphics[width=2.cm,height=3.4cm]{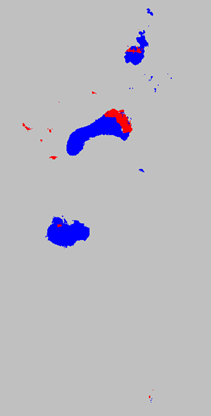} & \includegraphics[width=2.cm,height=3.4cm]{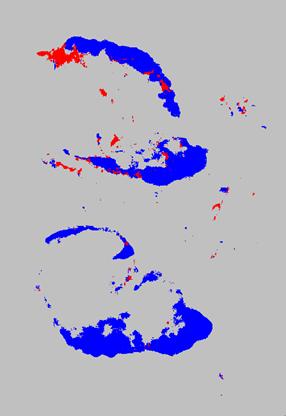}\\

\midrule
\rotatebox[origin=c]{270}{E.b} & 
\includegraphics[width=2.cm,height=3.4cm]{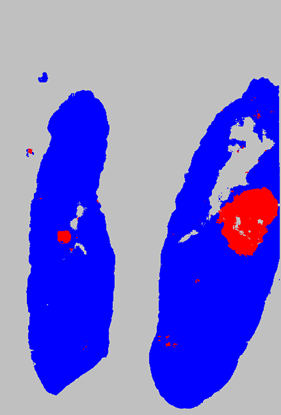} & \includegraphics[width=2.cm,height=3.4cm]{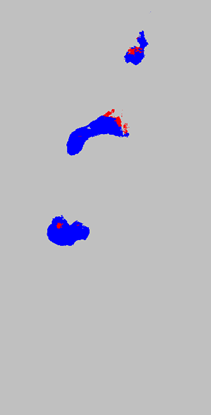} & \includegraphics[width=2.cm,height=3.4cm]{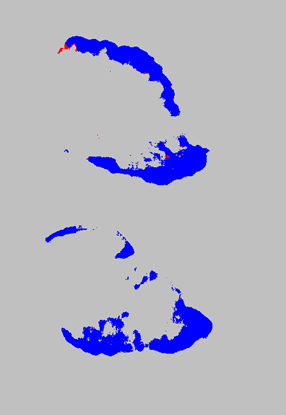}\\

\bottomrule
\end{tabular}
\label{tab:example-pred} 
\end{table}

\begin{figure*}[!t]
\centering
\includegraphics[width=0.9\textwidth]{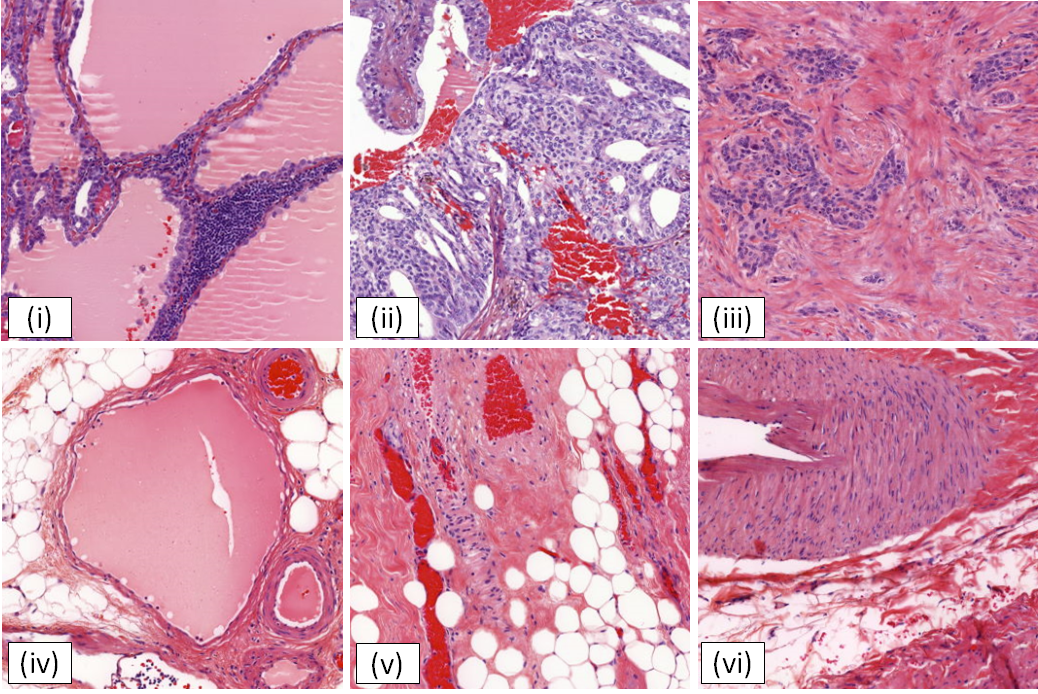}
\caption{Some examples of confusion between classes  $\mathbb{O}$ and $\mathbb{C}$ caused by inter-class similar region with common histological structures. Images (i), (ii) and (iii) represent patches from the training set that belong to category $\mathbb{C}$, which also contain fluid, blood and fibrosis with lymphocytes, respectively. Images (iv), (v) and (vi) represent patches from the test set that belong to category $\mathbb{O}$, containing the same histological structures as the above images. During training, if there are not enough examples of class $\mathbb{O}$ containing the mentioned structures, the respective histological structures in the test patches are predicted as class $\mathbb{C}$ (false positive). Best viewed in color.} 
\label{fig:conf1}
\end{figure*}

We further investigate the reasons why using the natural distribution turned out to be superior, by manually looking at the predicted masks. According to our thorough observation, we figured out that the miss-classification by E.a (balanced distribution) is due to the inter-class similar regions: the regions containing common histological structures, e.g., blood, fibrous tissue, etc. These histological structures are common in both ROI ($\mathbb{C}$ and $\lnot\mathbb{C}$) and non-ROI ($\mathbb{O}$) regions (see Fig.~\ref{fig:conf1}). When an ROI contains such a histological structure by comprising a small area in a gigapixel image, it is practical to overlook that small area and annotate it as a corresponding ROI ($\mathbb{C}$ or $\lnot\mathbb{C}$) instead of annotating it as non-ROI, although the annotation is actually wrong. 
During training, we thus need enough examples of these inter-class similar regions (i.e., over-representation of class $\mathbb{O}$) with their actual annotation to compensate the unavoidable pitfall of the annotation, otherwise it might cause false positives during prediction \cite{FP_shanks2013false}. 
 
\section{CONCLUSION}
\label{sec:conclusion}

In this research, we performed a data-level comparison between naturally imbalanced and artificially balanced distributions, thus determining the usability of the non-ROIs in WSIs for CNN training. According to our analysis, the commonly-prescribed balanced distribution for CNNs does not provide the best performance. Instead, the original training set, which is naturally biased towards the non-ROI class, produces the best performance for cancer detection {regarding less FPs generation yet maintaining a comparable or higher sensitivity or recall}. Moreover, the results illustrate the usefulness of the usually filtered out non-ROI data. Specifically, these examples are useful to be added during CNN training to solve the inter-class similarity between ROIs and non-ROIs, thus reducing false positives during prediction. Notably, prediction with less false positives reduces pathologists' workloads during cancer detection.  
We believe that the findings of this study can be useful for researchers in cancer detection from WSIs, providing a new viewpoint that can make them reconsider using the natural distribution and the non-ROIs in building training data. In some regard, our work is limited because we only considered the natural and the balanced distribution. In future work, it would be interesting to observe the impact of other distributions, e.g., lymph-node biased or cancer biased distributions. 
Furthermore, the usability of the mixed patches (where cancer and lymph-node pixels are both present), which are not explored yet, would be another compelling track to investigate. 

\section*{acknowledgements}
Authors would like to thank Professor Pierre Brousset, Dr. Camille Franchet, and Dr. Margot Gaspard for their fruitful discussions and providing the data set. \\
The research leading to these results has received funding from the NO Grants 2014-2021, under project ELO-Hyp contract no. 24/2020.

\bibliographystyle{ACM-Reference-Format}
\bibliography{references} 

\end{document}